\documentclass[10pt,twocolumn,letterpaper]{article}

\usepackage[pagenumbers]{cvpr} % To force page numbers, e.g. for an arXiv version

\usepackage{multirow}

\newcommand{\fns}{\footnotesize}
\definecolor{cvprblue}{rgb}{0.21,0.49,0.74}
\usepackage[pagebackref,breaklinks,colorlinks,allcolors=cvprblue]{hyperref}

\def\paperID{12506} % *** Enter the Paper ID here
\def\confName{CVPR}
\def\confYear{2025}

\title{PSA-SSL: Pose and Size-aware Self-Supervised Learning on LiDAR Point Clouds}

\author{Barza Nisar, Steven L. Waslander\\
University of Toronto Robotics Institute
}

\begin{document}
\maketitle
\begin{abstract}
   Self-supervised learning (SSL) on 3D point clouds has the potential to learn feature representations that can transfer to diverse sensors and multiple downstream perception tasks. However, recent SSL approaches fail to define pretext tasks that retain geometric information such as object pose and scale,  which can be detrimental to the performance of downstream localization and geometry-sensitive 3D scene understanding tasks, such as 3D semantic segmentation and 3D object detection. 
   We propose PSA-SSL, a novel extension to point cloud SSL that learns object pose and size-aware (PSA) features. Our approach defines a self-supervised bounding box regression pretext task, which retains object pose and size information. Furthermore, we incorporate LiDAR beam pattern augmentation on input point clouds, which encourages learning sensor-agnostic features. 
   Our experiments demonstrate that with a single pretrained model, our light-weight yet effective extensions achieve significant improvements on 3D semantic segmentation with limited labels across popular autonomous driving datasets (Waymo, nuScenes, SemanticKITTI). Moreover, our approach outperforms other state-of-the-art SSL methods on 3D semantic segmentation (using up to 10 times less labels), as well as on 3D object detection.  Our code will be released on https://github.com/TRAILab/PSA-SSL.
\end{abstract}

\section{Introduction}
\label{sec:intro}

\begin{figure}[tb]
  \centering
  \includegraphics[width=\columnwidth]{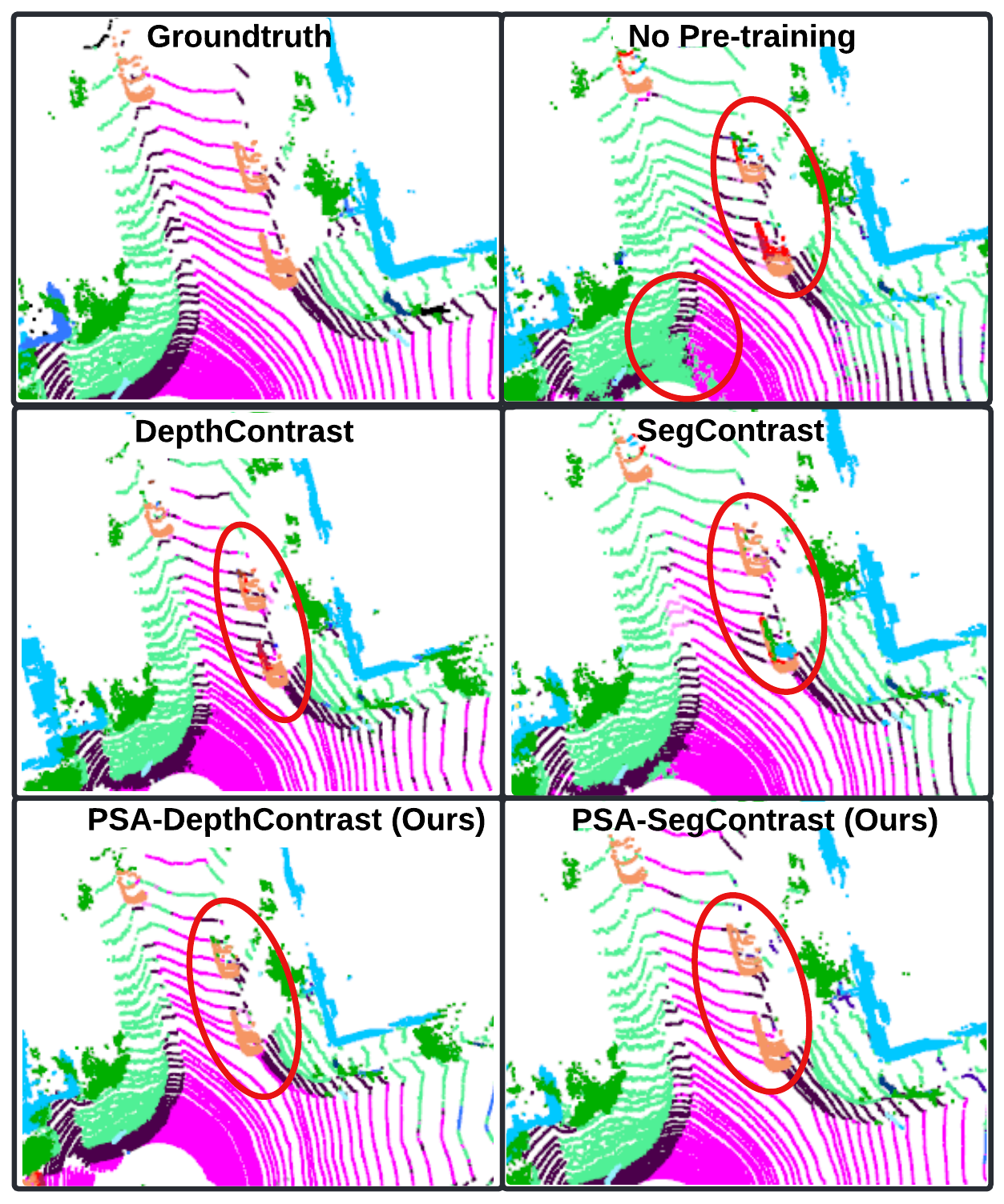}
  \vspace{-20pt}
  \caption{Comparison of qualitative semantic segmentation results of PSA-DepthContrast and PSA-SegContrast against their original baselines  \cite{Zhang_2021_depthcontrast, nunes2022segcontrast}  on SemanticKITTI validation scan. Our approach can capture the full extent of objects in the scene, thus exhibiting the least label confusion within a single object.}
  \vspace{-10pt}
  \label{fig:qualitative1}
\end{figure}

Autonomous vehicles often rely on LiDAR to obtain accurate 3D geometry and location of objects in the scene. LiDAR-based scene perception tasks, such as 3D semantic segmentation and 3D object detection, require learning from large labelled datasets. However, annotating 3D point cloud scenes is both time-consuming and expensive. Recently, self-supervised learning (SSL) on point clouds has gained popularity as useful and \emph{transferable} representations can be learned directly from large unlabeled datasets \cite{Zhang_2021_depthcontrast, nunes2022segcontrast, xie2020pointcontrast, yin2022proposalcontrast, sautier2024bevcontrast}, obviating the need for extensive labelling efforts when changing domains. 

Learning transferable representations on point clouds necessitates designing a single SSL framework that can learn features generalizable across various downstream tasks and LiDAR sensors.
A commonality among 3D perception tasks, such as semantic segmentation and object detection, is that they are object pose and geometry sensitive. However, recent SSL approaches for point clouds which employ contrastive learning (CL) \cite{Zhang_2021_depthcontrast, nunes2022segcontrast, tarl, sautier2024bevcontrast} fail to encode such critical geometric information. CL maximizes the similarity between differently rotated, translated or scaled versions of the same point cloud instance. The point cloud instance can be a single point, voxel, region, object or an entire scene. The learned features, thereby, become invariant to these geometric transformations, losing information about the object pose and size \cite{shi2022self}. We hypothesize that learning pose and geometry-insensitive features limits usefulness and generalizability to the various downstream 3D scene understanding tasks. 
Similarly, recent SSL approaches lack transferability to different LiDAR sensors. Most CL-based methods employ random dropping of points and cropping of cuboids from the LiDAR scan as additional augmentation strategies, which are not representative of different LiDAR beam patterns. Hence, their cross-LiDAR generalizability is limited. 

In this work, we aim to design a single CL-based SSL framework that can achieve both sensor and task generalizability. To this end, we propose PSA-SSL, a novel extension to CL-based point cloud SSL that learns LiDAR pattern agnostic and object pose and size-aware (PSA) features. 
In particular, we define 3D bounding box regression as a self-supervised pretext task which complements the CL task to retain object pose and size information. 
Although bounding box regression is a common task employed in supervised learning and self-training for 3D object detection \cite{pointrcnn, yan2018second, oyster}, to the best of our knowledge, we are the first to identify its usefulness as a self-supervised pretraining task which can solve a fundamental limitation of contrastive-based SSL approaches. \Cref{fig:qualitative1} qualitatively illustrates the consequence of this limitation, showing that the recent contrastive-based SSL approaches are more prone to label confusion within objects. 
Since our method additionally encodes object pose and size information, it learns features that can better capture the full extent of objects in the scenes, yielding reduced label mixup within a single object. 
Furthermore, we incorporate LiDAR beam pattern augmentation (LPA) \cite{lidomaug} during the augmentation stage of contrastive learning to learn sensor-agnostic features. 
As contrastive loss maximizes the similarity of embeddings from different beam patterns, the model learns representations invariant to various LiDAR sparsity patterns.
The LiDAR beam pattern augmentation (LPA) alleviates the need to pretrain on different LiDAR datasets, allowing a single pretrained model to transfer to various downstream LiDAR sensors. While LPA is not novel and has been adopted in domain adaptation methods \cite{hu2023density, yan2023spot, lidomaug}, we found that leveraging LPA in point cloud SSL significantly boosts performance on new LiDAR sensors. 

Our proposed framework is model-agnostic and can be applied to any CL or similarity loss-based SSL method for LiDAR. Moreover, our approach does not increase the pretraining times. In fact, on our setup, our approach exhibits improved performance at 33\% reduced pretraining time. 
Through our experiments, we show that our simple and light-weight PSA-SSL significantly improves performance on 3D semantic segmentation in low label regimes across different popular autonomous driving datasets (Waymo, nuScenes and SemanticKITTI). Our method also achieves SOTA performance in both 3D semantic segmentation and 3D object detection and demonstrates better cross-LiDAR generalizability.

While most CL literature for point clouds \cite{xie2020pointcontrast, Zhang_2021_depthcontrast, nunes2022segcontrast, yin2022proposalcontrast, sautier2024bevcontrast}  modifies the granularity of the features to improve results, our paper encourages readers to question the direction of this research by identifying a crucial limitation of contrastive learning on point cloud scenes. Our paper suggests practical solutions that can improve existing methods significantly while moving away from contrastive-only SSL. Our key contributions are summarized as follows:
\begin{itemize}
    \item We address a fundamental limitation of contrastive-based SSL methods for LiDAR point clouds by introducing bounding box regression as an SSL pretext task. 
    
    \item Our experiments show that learning object pose and size-aware features can significantly improve the performance of similarity-based SSL approaches on 3D semantic segmentation with limited labels, as well as outperform the SOTA methods. 

    \item 
    Our model pretrained on a single dataset shows improved performance on different popular autonomous driving datasets compared to existing methods on both 3D semantic segmentation and 3D object detection.
\end{itemize}

\section{Related Work}
\label{sec:related_work}
\subsection{Self-supervised Learning on Point Clouds}
SSL for LiDAR point clouds can be categorized into 4 families \cite{yan2024forging}: 1) Contrastive-based 2) Reconstruction-based \cite{ALSO, occmae, maeli} 3) Image-to-LiDAR Distillation-based \cite{slidr, stslidr, seal}, and 4) Rendering-based \cite{unipad}.
Distillation and rendering-based methods rely on image data for training, and therefore, require accurate camera-LiDAR sensor calibration. Most LiDAR-only approaches are either reconstruction or contrastive-based. Both families have been shown to perform competitively on scene perception tasks \cite{sautier2024bevcontrast}. Since our work aims to improve contrastive-based approaches, in this paper, we focus only on this family of SSL and present its related work in more detail.

\subsection{Self-supervised Contrastive Learning on Point Clouds}
Self-supervised contrastive learning literature for point clouds can be broadly divided into four categories in terms of the granularity of features contrasted: \textbf{1) Scene-level:} Scene-level approaches, such as DepthContrast \cite{Zhang_2021_depthcontrast}, construct a single feature vector for each point cloud scene and apply InfoNCE loss \cite{infonce} on these scene-wise features. \textbf{2) Point/Voxel-level:} PointContrast \cite{xie2020pointcontrast} applies InfoNCE loss over point-wise features sampled from differently augmented overlapping views of the same point cloud. 
A major limitation of point-level methods is that they can assign points belonging to the same object as negative examples in contrastive loss, which results in learning dissimilar features for semantically-similar points. To overcome this limitation, GCC-3D \cite{liang2021exploring} introduces a contrastive objective that enforces the spatially close voxels to have high feature similarity. 
\textbf{3) BEV-level:} BEVContrast \cite{sautier2024bevcontrast} contrast features at the level of 2D cells in the Bird’s Eye View plane. The method can compete with the performance of segment-level representations. However, it still suffers from the limitation of false negative sampling like point/voxel-level methods. \textbf{4) Region/Object-level:} SegContrast \cite{nunes2022segcontrast} exploits DBSCAN \cite{dbscan} to cluster non-ground points and applies InfoNCE loss to contrast cluster-wise features. TARL \cite{tarl} builds on SegContrast and extracts temporal views as augmented versions of the same object. ProposalContrast \cite{yin2022proposalcontrast} adopts both contrastive learning and SwAV \cite{caron2020unsupervised} to learn instance and class-discriminative features on regions.
Point features of each region are aggregated by attention, which encodes geometrical relations among the points. 

While scene-level methods struggle to capture local geometric structures of road objects, point/voxel-level methods over-emphasize fine-grained details, lacking the receptive field to describe object-level characteristics. Region/Object-level methods \cite{nunes2022segcontrast, yin2022proposalcontrast} extract and contrast region or object-level instances, which can benefit both 3D segmentation and 3D object detection. However, a fundamental issue with most SSL methods is that they use an InfoNCE loss alone, which encourages feature similarity across different geometric transformations (i.e. different rotation, translation and scaling) of the same region, leading to loss of information about object pose and size. While Shi and Rajkumar \cite{shi2022self} attempt to combine contrastive loss with geometric pretext tasks, the authors only introduce rotation and scale prediction pretext tasks and contrast point-level representations, which have limited receptive field for learning useful representations. Moreover, they evaluate their method only on object detection and not on semantic segmentation. In our work, we propose pretraining the point cloud feature extractor jointly with a bounding box regression task and a contrastive loss, which can help learn class, pose and size discriminative representations. Our experiments show performance improvement on both 3D semantic segmentation and object detection. 
Another main limitation of the aforementioned SSL methods is that they usually pretrain and fine-tune on the same dataset and have limited results on transfer learning to different LiDAR sensors. 

\subsection{LiDAR Beam Pattern Augmentation}

LiDAR beam augmentation has been commonly adopted in domain adaptation literature to learn beam density-invariant features \cite{hu2023density}. 
Moving away from unrealistic random LiDAR beam resampling \cite{hu2023density, yan2023spot}, Ryu \etal~\cite{lidomaug} proposed LiDomAug, which is a LiDAR beam pattern augmentation strategy that can transform the input point cloud with arbitrary cylindrical LiDAR configurations, mounting pose and entangled motions of LiDAR spin and moving platform. 
Since existing SSL approaches for LiDAR point clouds lack cross-sensor generalizability, our work incorporates LiDAR beam pattern transformation as an augmentation strategy to learn sensor-agnostic representations in a self-supervised contrastive learning framework. 

\section{Method}
\subsection{Pipeline Overview}
\begin{figure*}[tb]
  \centering
  \includegraphics[width=0.9\textwidth]{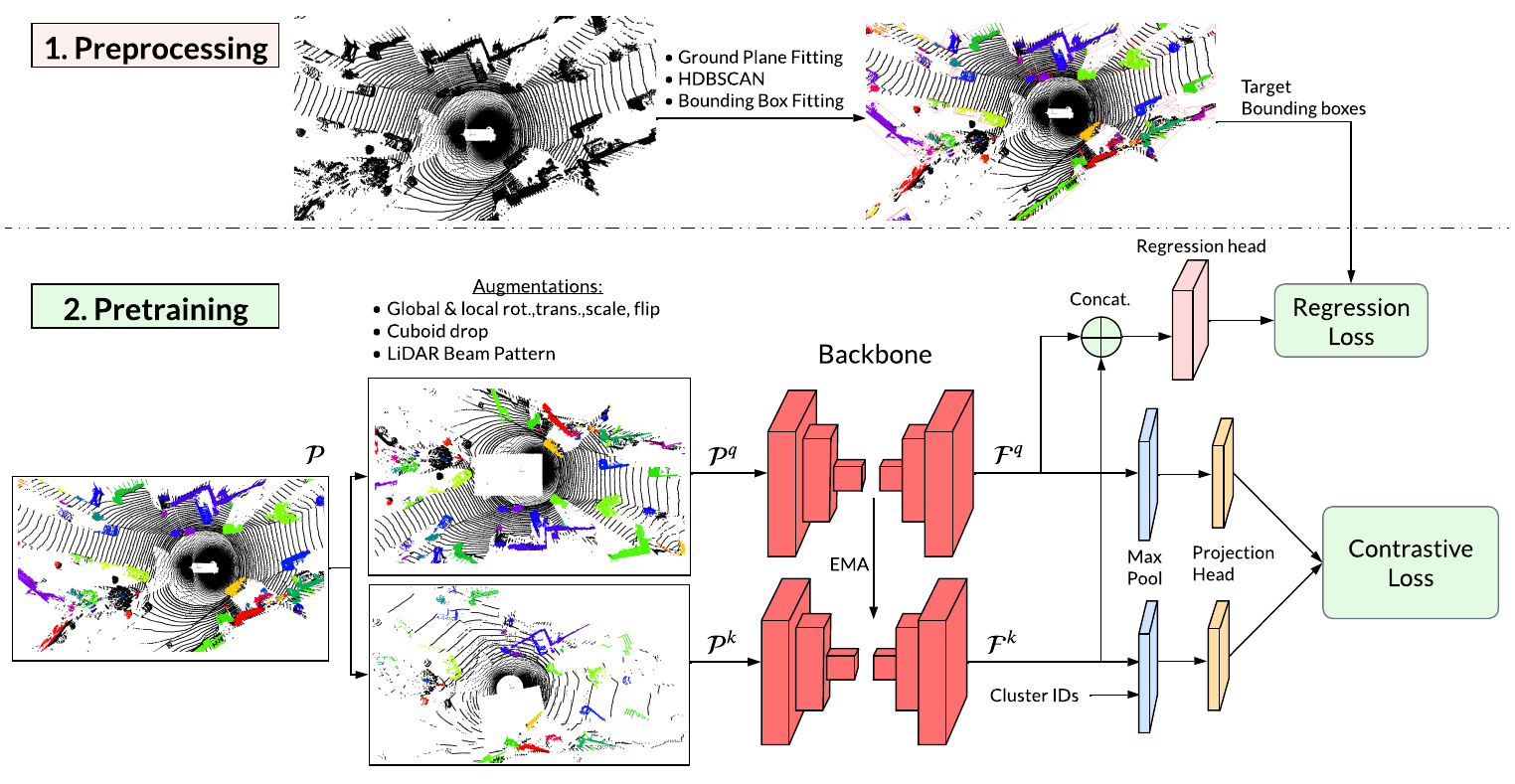}
  \vspace{-10pt}
  \caption{An overview of our self-supervised point cloud representation learning framework.
  }
  \vspace{-10pt}
  \label{fig:shape-ssl-overview}
\end{figure*}
\Cref{fig:shape-ssl-overview} presents an overview of our framework. Our method involves three stages: 1) Preprocessing, 2) Pretraining 3) Fine-tuning. In the preprocessing stage, we fit bounding boxes on clustered LiDAR scans, which serve as targets for the bounding box regression pretext task. The pretraining stage trains a point cloud feature extractor with the contrastive objective and bounding box regression loss jointly. For the contrastive loss, we generate two views of each scan using LiDAR beam pattern transformation and standard data augmentations such as random rotation, translation, scaling and cuboid drops. After pretraining, a semantic segmentation or an object detection head is added to the pretrained backbone and the entire model is fine-tuned on the new task and/or dataset. The next sections will explain each stage in detail.

\subsubsection{Preprocessing} 
In the preprocessing stage, we generate the bounding box targets to supervise the regression task. To this end, we first use Patchwork++ \cite{patchworkpp}, with its default parameters, to fit a ground plane and partition the point cloud into ground and non-ground points. We then use a hierarchical density-based spatial clustering algorithm, HDBSCAN \cite{hdbscan} to cluster the non-ground points and assign cluster ID to each point.  
To avoid over-clustering, we reject clusters with points less than a minimum cluster size. We use the default value of minimum cluster size from SegContrast \cite{nunes2022segcontrast} (i.e. 20 points) and do not tune this for any of our pretraining dataset. The clustering $\epsilon$ in HDBSCAN controls the over/under segmentation, and we tune it by visually inspecting the clustering on a small number of frames.
Following the \textit{common-sense} heuristics proposed in \cite{modest}, we also reject clusters that are floating in the air, are under the ground plane or have an exceptionally large volume. The parameters of these heuristics are taken from \cite{modest} and are not tuned for any of our pretraining datasets. 
We assume that each cluster represents an individual object in the scene. Finally, an off-the-shelf bounding box fitting algorithm \cite{lshape} is used to fit upright bounding boxes around the clusters.

\subsubsection{Pretraining} 
Given a clustered point cloud $\mathcal{P} = \{\mathbf{p}_1, \hdots, \mathbf{p}_N\}$ with $|\mathcal{P}| = N$ points, where $\mathbf{p}_i \in \mathbb{R}^{(3+f_p)}$ is a vector of 3D point coordinates and $f_p$ point features (e.g. intensity), we first generate two views $\mathcal{P}^{q}$ and $\mathcal{P}^{k}$ of the point cloud by subjecting it to several data augmentation strategies, such as LiDAR beam pattern transformation, random cuboid dropouts, random flip, rotation, translation and scaling. Since these point cloud transformations change the pose and size of the object clusters, we also transform the target bounding boxes with the same augmentations. The augmented pair of point clouds is then processed by a 3D point cloud encoder $f_\theta: \mathbb{R}^{N \times (3+f_p)} \rightarrow \mathbb{R}^{N \times D}$ which extracts point-wise features $\mathcal{F}^{q}$ and $\mathcal{F}^{k}$. 
More specifically, following the MoCo \cite{he2020momentum} framework, $\mathcal{P}^{q}$ is processed by the query encoder $f_{\theta_q}$ and $\mathcal{P}^{k}$ is processed by the momentum encoder $f_{\theta_k}$. 
Point-wise features are then processed by two parallel heads: 1) Regression head: which predicts bounding boxes on the concatenated $\mathcal{F}^{q}$ and $\mathcal{F}^{k}$. 2) Contrastive head: Depending on the baseline SSL method, max-pooling is performed on point features to either obtain cluster-level features (for SegContrast) or scene-level features (for DepthContrast). These features are then passed through a projection head and subsequently used to compute contrastive loss.
For details on the contrastive loss formulation, refer to \cite{nunes2022segcontrast, Zhang_2021_depthcontrast}.

\subsection{LiDAR Beam Pattern Augmentation}
Existing autonomous driving datasets use LiDAR with different fields-of-view, resolutions, channels and sensor positions. To learn representations that can generalize to different LiDARs, we transform the input point cloud using randomized LiDAR configurations during the data augmentation stage \cite{lidomaug}. The configuration of a LiDAR can be defined by its vertical field-of-view $(f_{down}, f_{up})$ and the resolution of the range map $(H, W)$. 
We randomly select a LiDAR configuration from several predefined configurations and express the point cloud as a range map via spherical projection $\Pi(\mathbf{x})$ of 3D points. The spherical projection is computed using $H, W, f_{down}, f_{up}$ (refer to \cite{lidomaug} for the projection equation).
The range map is then transformed back to the point cloud using the inverse of the projection model $\Pi^{-1}$. 
We apply the same transformations to the target bounding box center coordinates so that they align with the transformed point cloud. In our implementation, we transform one of the two views using a randomly sampled LiDAR configuration. We also experiment with transforming the point cloud into multiple different LiDAR configurations and mixing the resulting point clouds along randomly sampled azimuth angle ranges, similar to PolarMix \cite{polarmix}. We perform ablations on the two variations, and refer to them as 1) Single Pattern 2) PolarMix respectively.
When contrastive loss maximises the similarity of embeddings from different beam patterns, the model learns representations invariant to various LiDAR sparsity patterns. 

\subsection{Bounding Box Regression Head} 
Since our data augmentations include random rotations, translation, and scaling of clusters in the point cloud, contrastive objective will learn representations invariant to these geometric transformations. If the backbone is voxel-based, the voxel grid can help recover the 3D positions and scale of the cluster features, resulting in representations equivariant to translation and scaling. If the backbone is PointNet-based, we need to store the 3D point coordinates in memory and append these to the features extracted to retain the position information. However, the learned representations alone lose the position, scale and rotation information as the contrastive loss enforces similarity between different geometrically transformed versions of the same clusters. 

To encode geometric information, such as object pose and size, alongside class discriminative features, we propose adding a bounding box regression head to the concatenated outputs of the query and momentum feature encoders. Our regression network is simply a fully-connected neural network (with two hidden layers of dimension 256 neurons each) that predicts for each cluster point the bounding box offset from a fixed-size anchor, similar to the anchor-based 3D object detectors \cite{pointrcnn, yan2018second}. We compute a smooth L1 loss \cite{fastrcnn} between the predicted and target regression offsets only for the clustered points. While the contrastive loss learns class-discriminative features, the regression head retains the geometric content of each object. Our pretraining loss is a linear combination of contrastive and regression loss $\mathcal{L} = \beta_1\mathcal{L}_{con} + \beta_2\mathcal{L}_{reg}$.

\section{Implementation and Experimental Setup}
\label{sec:exp}

\textbf{Baselines:} We apply the regression pretext task and augmentation strategy to DepthContrast (DC) \cite{Zhang_2021_depthcontrast} and SegContrast (SC) \cite{nunes2022segcontrast} and compare the performance against their original method. DC learns scene-level representations that are invariant to various geometric transformations and to point-voxel encoding. For invariance to the latter, DC requires storing additional voxel backbones and its negative feature bank in memory, significantly limiting the batch size possible per GPU. To use the same batch size as other baselines, we construct scene-level loss only on the point encoding of the scene (\ie `within format' loss \cite{Zhang_2021_depthcontrast}) to replicate the DepthContrast baseline. Additionally, we implement two more baselines similar to ProposalContrast \cite{yin2022proposalcontrast} and \cite{shi2022self} since these methods are closely related to our work in terms of learning geometric features in addition to contrastive learning. 
In implementing ProposalContrast, we add their attention-based proposal encoding module to encode cluster features, instead of encoding fixed-sized regions selected by farthest point sampling. 
For implementing \cite{shi2022self}, we add two regression heads atop SegContrast to predict 1) relative scales and 2) heading difference between the two views of each cluster in the batch.  
For all our experiments we use the MinkUNet \cite{minkowski} backbone.  

\textbf{Pretraining datasets:} We pretrain 3D backbones of all models on 10\% of the train split of the Waymo Open Dataset (Waymo) \cite{waymo}. Additionally, to show that our method is robust to the dataset preprocessing parameters and to compare against SOTA on common benchmarks, we also pretrain on nuScenes \cite{nuscenes} and SemanticKITTI \cite{behley2019semantickitti} using the same preprocessing parameters as Waymo, except for the clustering epsilon. We pretrain on nuScenes and SemanticKITTI using the same train splits as in \cite{sautier2024bevcontrast, ALSO} for a fair comparison.

\textbf{Pretraining parameters:} To generate target bounding box regression offsets, we use a cube-shaped anchor of dimension 1 on each point. We set the values for $\beta_1=1$ and $\beta_2=0.5$, which are the contrastive and regression loss weights respectively. For both `Single Pattern' and `PolarMix' LiDAR pattern augmentations, we randomly select LiDAR configurations `Velodyne-32 (v32)', `Velodyne-64 (v64)' and `Ouster-64 (o64)' with probabilities [0.6, 0.2, 0.2] respectively. Section 6.1.1 in supplementary material presents a sensitivity analysis to these probabilities. We refer the reader to \cite{lidomaug} for the specifications (channels, range, field-of-view, horizontal angular resolution) of `v32',`v64',`o64' LiDARs. For PolarMix, we render the point cloud into a randomly chosen configuration 5 times and, each time, uniformly sample the azimuth angle crop range between 10 to 90 degrees. Following the pretraining scheme in \cite{Zhang_2021_depthcontrast, nunes2022segcontrast}, we use the stochastic gradient descent (SGD) optimizer with a momentum of 0.9, learning rate of 0.12 and weight decay of 0.0001. 
We set the contrastive loss temperature $\tau=0.1$ and $\tau=0.04$ for DC and SC, respectively. We use a batch size of 32, 32, 64 to pretrain on Waymo, nuScenes and SemanticKITTI, respectively and use 2 NVIDIA RTX 6000 Ada 48 GB GPUs. All models are pretrained for 200 epochs, except for the sensitivity analysis experiments, which are pretrained for 30 epochs.

\textbf{Preprocessing Hyperparameters:} 
We set the clustering $\epsilon$ for Waymo, nuScenes and SemanticKITTI to 0.2, 0.3 and 0.25 respectively. 
Since clustering $\epsilon$ is the only preprocessing hyperparameter we tuned for the pretraining datasets, we provide a sensitivity analysis to this parameter in Sec. 6.1.2 of the supplementary materials.

\textbf{Fine-tuning:} To evaluate the quality of the SSL representations, we fine-tune the pretrained models on 3D semantic segmentation and 3D object detection. All segmentation and detection results are reported on the validation splits of the respective datasets. We compare our method against baseline models which are pretrained and fine-tuned using the same training and preprocessing parameters for a fair comparison (see \cref{tab:semseg_main}, \cref{tab:3dod}). We further show our method's performance compared to SOTA published results using common fine-tuning protocols and data splits (see \cref{tab:nusseg}, \cref{tab:kittiseg}, \cref{tab:tarlseg}, \cref{tab:kitti3dod}).

\textbf{1. 3D Semantic Segmentation comparison against baselines:} For models pretrained on Waymo, we fine-tune them on Waymo, nuScenes, and SemanticKITTI \cite{behley2019semantickitti} datasets. The point cloud pattern differs significantly in all three datasets, demonstrating our method's ability to generalize across sensor types. We use the same optimizer parameters for segmentation experiments as in the pretraining but with a learning rate of 0.24. To show the effectiveness of SSL when labels are limited, existing approaches \cite{nunes2022segcontrast, Zhang_2021_depthcontrast} typically fine-tune on low label regimes. Hence, following their standard, we fine-tune on 1\% and 5\% of each dataset's annotated train split for 100 and 50 epochs, respectively, with a batch size of 16. 
To create 1\% and 5\% splits, we sample the train splits by setting a fixed frame sampling interval. Hence, all experiments use the same subset of training data for each percentage. 
Segmentation results (mIoU) are reported as the average and standard deviation of 3 fine-tuning runs in \cref{tab:semseg_main}. 
We use the standard training and validation splits for all the datasets and validation splits are not used in the pretraining or fine-tuning. For SemanticKITTI, we use sequences 0 to 10 (except 8) for training and sequence 8 for validation.   

\textbf{2. 3D Semantic Segmentation comparison against SOTA on common benchmarks:} To show a fair comparison of our model's performance against SOTA, we take the models pretrained on nuScenes and SemanticKITTI and follow the same fine-tuning protocol (splits, batch size, epochs, optimizer) as in \cite{sautier2024bevcontrast, ALSO}. The results (mIoU) are reported as the average and standard deviation over 5 fine-tuning runs in \cref{tab:nusseg} and \cref{tab:kittiseg}. We further compare the cross-LiDAR generalizability in \cref{tab:tarlseg} using the same fine-tuning protocol as in \cite{tarl}. 

\textbf{3. 3D Object Detection comparison against baselines:} We take our model pretrained on Waymo and fine-tune on 1\% Waymo, 5\% nuScenes and 5\% KITTI \cite{kitti} datasets. For KITTI, we use the same train splits as in \cite{Zhang_2021_depthcontrast}. Following \cite{nunes2022segcontrast, tarl}, we employ PartA2 \cite{parta2} as the base detector, with the same hyperparameters as in \cite{nunes2022segcontrast, tarl} (batch size of 2). Since the PartA2 detection head is larger than the MinkUNet backbone (the head has 57.5M while the backbone has 21.7M parameters), we first warm up the detection head on the frozen pretrained backbone for 5 epochs and then fine-tune the whole model for 80 epochs. The results are reported in \cref{tab:3dod}.

\textbf{4. 3D Object Detection comparison against SOTA on common benchmarks:} We take our model pretrained on SemanticKITTI and fine-tune on full KITTI train set using the same fine-tuning protocol as in \cite{tarl}. 3D AP is reported with 40 recall positions (R$_{40}$) and IoU thresholds of 0.7, 0.5, 0.5 for Car, Pedestrian and Cyclist respectively (AP$_{50}$) in \cref{tab:kitti3dod}.

\section{Results}
Our experiments show that PSA-SSL 1) significantly improves the recent contrastive learning approaches and outperforms SOTA SSL methods on 3D semantic segmentation and 3D object detection, 
2) outperforms existing methods on both tasks when transferred to different LiDAR sensors, 
3) does not increase the pretraining time.

\subsection{3D Semantic Segmentation}

\begin{table*}[t]
\centering
\resizebox{\textwidth}{!}{%
\begin{tabular}{c|ccc|ccc}
\toprule
& \multicolumn{3}{c|}{1\% labels} & \multicolumn{3}{c}{5\% labels} \\  
Method & Waymo & nuScenes & SemKITTI & Waymo & nuScenes & SemKITTI \\ 
\midrule
No pretraining & 49.34 \, \fns{$\pm$ 0.78} & 35.71 \, \fns{$\pm$ 0.12} & 41.95 \, \fns{$\pm$ 0.58} & 59.13 \, \fns{$\pm$ 0.37} & 50.83 \, \fns{$\pm$ 0.47} & 56.05 \, \fns{$\pm$ 0.45} \\ \hline

DC \cite{Zhang_2021_depthcontrast} & 50.36 \, \fns{$\pm$ 0.38} & 35.32 \, \fns{$\pm$ 0.13} & 48.77 \, \fns{$\pm$ 0.28} & 58.97 \, \fns{$\pm$ 0.10} & 51.16 \, \fns{$\pm$ 0.24} & 58.09 \, \fns{$\pm$ 0.35} \\ 

PSA-DC (Ours) & \textbf{53.02} \, \fns{$\pm$ 0.05} & \textbf{37.75} \, \fns{$\pm$ 0.33} & \textbf{49.92} \, \fns{$\pm$ 0.96} & \textbf{61.25} \, \fns{$\pm$ 0.22} & \textbf{51.78} \, \fns{$\pm$ 0.15} & \textbf{58.66} \, \fns{$\pm$ 0.44} \\ \hline

\rowcolor{lime}
\textit{Improvement} & +2.66 & +2.43 & +1.15 & +2.28 & +0.63 & +0.57 \\ \hline

SC \cite{nunes2022segcontrast} & 53.50 \, \fns{$\pm$ 0.13} & 36.01 \, \fns{$\pm$ 0.51} & 49.72 \, \fns{$\pm$ 0.83} & 61.24 \, \fns{$\pm$ 0.21} & 51.79 \, \fns{$\pm$ 0.16} & 58.17 \, \fns{$\pm$ 0.42} \\ 

SC + Attn. Encoder\cite{yin2022proposalcontrast} & 52.37 \, \fns{$\pm$ 0.40} & 36.13 \, \fns{$\pm$ 0.58} & 47.73 \, \fns{$\pm$ 1.13} & 60.74 \, \fns{$\pm$ 0.23} & 50.63 \, \fns{$\pm$ 0.22} & 58.02 \, \fns{$\pm$ 0.70} \\

SC + $\Delta$ Scale/Rot. Reg.\cite{shi2022self} & 52.98 \, \fns{$\pm$ 0.16} & 36.12 \, \fns{$\pm$ 0.47} & 49.61 \, \fns{$\pm$ 0.51} & 61.58 \, \fns{$\pm$ 0.05} & 51.53 \, \fns{$\pm$ 0.20} & 57.98 \, \fns{$\pm$ 0.30} \\ 
PSA-SC (Ours) & \textbf{54.36} \, \fns{$\pm$ 0.33} & \textbf{37.89} \, \fns{$\pm$ 0.27} & \textbf{52.11} \, \fns{$\pm$ 0.81} & \textbf{62.74} \, \fns{$\pm$ 0.13} & \textbf{52.54} \, \fns{$\pm$ 0.24} & \textbf{59.12} \, \fns{$\pm$ 0.16} \\ \hline
\rowcolor{lime}
\textit{Improvement} & +0.86 & +1.87 & +2.39 & +1.50 & +0.75 & +0.95 \\ 
\bottomrule
\end{tabular}
}
\vspace{-5pt}
\caption{Semantic segmentation performance (average and standard deviation in mIoU) for pretraining on Waymo and fine-tuning on 1\% and 5\% of labelled Waymo, nuScenes, SemanticKITTI.}
\vspace{-5pt}
\label{tab:semseg_main}
\end{table*}

\textbf{Comparison against baselines:} We evaluate the transferability of our method's learned representations on different downstream LiDAR sensors by fine-tuning on 3D semantic segmentation. 
\Cref{tab:semseg_main} presents the results for pretraining on Waymo and fine-tuning on different percentages of labelled training data and different datasets. As expected, all baseline SSL approaches, except DC, outperform training from a randomly initialized backbone. Since SC is a region-level method, it performs better than scene-level DC. We note that the performance of DC can be improved if pretrained with a much larger batch size or including `across format loss'(\ie \cite{Zhang_2021_depthcontrast} uses a batch size of 1024 and an additional point-to-voxel contrastive loss). Incorporating bounding box regression pretext task and LiDAR augmentation to DC (\ie PSA-DC), significantly improves performance across all datasets such that PSA-DC performs competitively compared to SC without needing to increase the batch size or including an additional voxel backbone that is memory intensive. We reason that since DC is a scene-level method, our bounding box regression pretext task enables learning of more fine-grained/object level features.
Adding an attention encoding module \cite{yin2022proposalcontrast} to SC or a regression head that predicts ratio of scales and rotation difference between the two views \cite{shi2022self} gives similar performance compared to the baseline SC. 
On the other hand, applying PSA to both baselines leads to statistically significant performance gains with the highest improvement seen on 1\% Waymo at \textbf{+2.66} mIoU for DC and on 1\% SemanticKITTI at \textbf{+2.39} mIoU for SC. 
Since our SSL framework was pretrained only on Waymo, considerable performance improvements on nuScenes and SemanticKITTI highlight the cross-LiDAR generalizability of our approach.

\textbf{Comparison against SOTA on common benchmarks:} \Cref{tab:nusseg} and \cref{tab:kittiseg} compares our method against SOTA SSL techniques, pretrained on nuScenes and SemanticKITTI respectively. Results for all SOTA methods are reported from \cite{sautier2024bevcontrast, maeli}. Since our LiDAR pattern augmentation (LPA) strategy only allows transformation from dense to sparse LiDARs, we omit its use when pretraining on nuScenes. 
The results show that PSA-SC significantly outperforms the SOTA contrastive learning \cite{sautier2024bevcontrast} and reconstruction-based method \cite{maeli} on all fine-tuning splits except on 0.1\%. Fine-tuned on only 10\% SemanticKITTI labels, our method reaches the performance of BEVContrast and MAELi trained on 100\% labels. Likewise, our method, pretrained on only 50\% nuScenes labels, is competitive with BEVContrast's performance using 100 \% labels. This reinforces our method's ability to generalize well with limited labels. \cref{tab:tarlseg} shows that our method has strong cross-LiDAR generalizability as it outperforms TARL \cite{tarl} by +3.56 mIoU when transferring from SemanticKITTI to nuScenes. 

\begin{table}[ht]
\centering
\resizebox{\columnwidth}{!}{%
\begin{tabular}{cccccc}
\toprule
Method &  0.1\% & 1\% &  10\% & 50\% & 100\% \\ 
\midrule
No pretraining & 21.6 \fns{$\pm$ 0.5} & 35.0  \fns{$\pm$ 0.3} & 57.3  \fns{$\pm$ 0.4} & 69.0  \fns{$\pm$ 0.2} & 71.2  \fns{$\pm$ 0.2} \\
PointContrast \cite{xie2020pointcontrast} &\textbf{27.1}  \fns{$\pm$ 0.5} & 37.0  \fns{$\pm$ 0.5} & 58.9  \fns{$\pm$ 0.2} & 69.4  \fns{$\pm$ 0.3} & 71.1  \fns{$\pm$ 0.2} \\
DepthContrast \cite{Zhang_2021_depthcontrast} & 21.7  \fns{$\pm$ 0.3} & 34.6  \fns{$\pm$ 0.5} & 57.4  \fns{$\pm$ 0.5} & 69.2  \fns{$\pm$ 0.3} & 71.2  \fns{$\pm$ 0.2}\\
ALSO \cite{ALSO} & 26.2  \fns{$\pm$ 0.5} & 37.4  \fns{$\pm$ 0.3} & 59.0  \fns{$\pm$ 0.4} & 69.8  \fns{$\pm$ 0.2} & 71.8  \fns{$\pm$ 0.2} \\
BEVContrast \cite{sautier2024bevcontrast} & 26.6  \fns{$\pm$ 0.5} & 37.9  \fns{$\pm$ 0.4} & 59.0  \fns{$\pm$ 0.6} & 70.5  \fns{$\pm$ 0.2} & 72.2  \fns{$\pm$ 0.1} \\
PSA-SC (Ours)$^\dagger$ & 25.1  \fns{$\pm$ 0.3} & \textbf{38.4}  \fns{$\pm$ 0.5} & \textbf{60.3}  \fns{$\pm$ 0.2} & \textbf{72.0}  \fns{$\pm$ 0.1} & \textbf{73.9}  \fns{$\pm$ 0.2} \\
\hline 
\end{tabular}%
}
\vspace{-5pt}
\caption{Semantic segmentation (average and standard deviation in mIoU) performance for pretraining on nuScenes and fine-tuning on different \% of the train split of nuScenes. $\dagger$ Without LPA.}
\label{tab:nusseg}
\end{table}

\begin{table}[ht]
\centering
\resizebox{\columnwidth}{!}{%
\begin{tabular}{cccccc}
\toprule
Method &  0.1\% & 1\% &  10\% & 50\% & 100\% \\ 
\midrule
No pretraining & 30.0 \fns{$\pm$ 0.2} & 46.2  \fns{$\pm$ 0.6} & 57.6  \fns{$\pm$ 0.9} & 61.8  \fns{$\pm$ 0.4} & 62.7  \fns{$\pm$ 0.3} \\
PointContrast \cite{xie2020pointcontrast} & 32.4  \fns{$\pm$ 0.5} & 47.9  \fns{$\pm$ 0.5} & 59.7 \fns{$\pm$ 0.5} & 62.7 \fns{$\pm$ 0.3} & 63.4 \fns{$\pm$ 0.4} \\
SegContrast \cite{nunes2022segcontrast} & 32.3 \fns{$\pm$ 0.3} & 48.9  \fns{$\pm$ 0.3} & 58.7 \fns{$\pm$ 0.5} & 62.1 \fns{$\pm$ 0.4} & 62.3 \fns{$\pm$ 0.4}\\
DepthContrast \cite{Zhang_2021_depthcontrast} & 32.5 \fns{$\pm$ 0.4} & 49.0  \fns{$\pm$ 0.4} & 60.3  \fns{$\pm$ 0.5} & 62.9  \fns{$\pm$ 0.5} & 63.9  \fns{$\pm$ 0.4}\\
ALSO \cite{ALSO} & 35.0 \fns{$\pm$ 0.1} & 50.0 \fns{$\pm$ 0.4} & 60.5  \fns{$\pm$ 0.1} & 63.4  \fns{$\pm$ 0.5} & 63.6  \fns{$\pm$ 0.5} \\
TARL \cite{tarl} & 37.9 \fns{$\pm$ 0.4} & 52.5 \fns{$\pm$ 0.5} & 61.2  \fns{$\pm$ 0.3} & 63.4  \fns{$\pm$ 0.2} & 63.7  \fns{$\pm$ 0.3} \\
MAELi \cite{maeli} & 34.6 & 50.7 & 61.3  & 63.6  & 64.2 \\
BEVContrast \cite{sautier2024bevcontrast} & \textbf{39.7}  \fns{$\pm$ 0.9} & 53.8  \fns{$\pm$ 1.0} & 61.4  \fns{$\pm$ 0.4} & 63.4  \fns{$\pm$ 0.6} & 64.1  \fns{$\pm$ 0.4} \\
PSA-SC (Ours) & 37.7 \fns{$\pm$ 0.1} & \textbf{55.3}  \fns{$\pm$ 0.2} & \textbf{64.2}  \fns{$\pm$ 0.4} & \textbf{64.7} \fns{$\pm$ 0.3} & \textbf{65.1}  \fns{$\pm$ 0.3} \\
\hline 
\end{tabular}%
}
\vspace{-5pt}
\caption{Semantic segmentation (average and standard deviation in mIoU) performance for pretraining on SemanticKITTI and fine-tuning on different \% of the train split of SemanticKITTI.}
\label{tab:kittiseg}
\end{table}

\begin{table}[ht]
\centering
\resizebox{0.6\columnwidth}{!}{%
\begin{tabular}{ccc}
\toprule
Method &  Mini & Full \\ 
\midrule
TARL \cite{tarl} & 39.36 & 68.26 \\
PSA-SC (Ours) & \textbf{39.60 (+0.24)} & \textbf{71.82 (+3.56)} \\
\hline 
\end{tabular}%
}
\vspace{-5pt}
\caption{Semantic segmentation mIOU for pretraining on SemanticKITTI and fine-tuning on nuScenes mini and full train sets.}
\label{tab:tarlseg}
\end{table}

\subsection{Object Detection}
\label{sec:3dod}

\begin{table*}[t]
\centering
\resizebox{\textwidth}{!}{%
\begin{tabular}{l|cccc|cccc|cc}
\toprule
\multirow{2}{*}{Method} & \multicolumn{4}{c|}{1\% Waymo (L1/L2 AP)} & \multicolumn{4}{c|}{5\% KITTI} 
& \multicolumn{2}{c}{5\% nuScenes} \\
& Vehicle & Pedestrian & Cyclist & mAP & Car & Ped. & Cyc. & mAP & mAP & NDS \\
\midrule

No pretraining  & 46.86/40.33 & 23.39/19.49 & 35.71/34.31 & 35.32/31.38 & 66.81 & \textbf{54.06} & 46.40 & 55.76 & 13.72 & 24.71 \\
SC \cite{nunes2022segcontrast} & 46.89/40.37 & \textbf{24.51/20.43} & \textbf{39.58/38.03} & \textbf{36.99/32.94} & \underline{68.33} & 51.05 & \underline{56.16} & \underline{58.51} 
& 14.03 & 26.00 \\
PSA-SC (Ours) & \textbf{48.02/41.4} & 24.07/20.07 & 37.36/35.89 & 36.48/32.45 & \textbf{68.86} & \underline{53.73} & \textbf{57.53} & \textbf{60.04} & \textbf{14.73} & \textbf{26.72}\\
\hline
\rowcolor{lime}
\textit{Improvement} & +1.13/+1.03 & -0.44/-0.36 & -2.22/-2.14 & -0.51/-0.49 & +0.53 & -0.33 & +1.36 & +1.53 & +0.60 & +0.72 \\ 

\bottomrule
\end{tabular}%
}
\vspace{-5pt}
\caption{3D object detection results for pretraining on Waymo and fine-tuning on 1\% Waymo and 5\% KITTI and 5\% nuScenes. For KITTI, 3D AP$_{50}$ at R$_{40}$ is reported for moderate difficulty level.}
\label{tab:3dod}
\end{table*}

\begin{table*}[ht]
\centering
\resizebox{0.7\textwidth}{!}{%
\begin{tabular}{c|c|ccc|ccc|ccc}
\toprule
 & & \multicolumn{3}{c}{Car} & \multicolumn{3}{c}{Pedestrian} & \multicolumn{3}{c}{Cyclist} \\ 
Method & mAP & E & M & H &  E & M & H &  E & M & H \\
\midrule
No pretraining & 72.66 & 91.38 & 81.91 & 79.95 & 67.68 & 61.90 & 56.42 & 93.19 & 74.18 & 70.83 \\
\midrule

SegContrast \cite{nunes2022segcontrast} & 73.52 & 92.08 & 82.10 & 80.18 & 68.72 & 63.77 & 57.36 & 91.26 & 74.70 & 70.28\\
TARL \cite{tarl} & 73.86 & \textbf{92.09} & 82.15 & 80.10 &  68.50 & 64.00 & \textbf{58.74} &  92.61 & 75.44 & 70.97\\
PSA-SC (Ours) &  \textbf{74.08} & 91.99 & \textbf{82.23} & \textbf{80.23} &  \textbf{69.44} & \textbf{64.18} & 57.51 &  \textbf{92.99} & \textbf{75.82} & \textbf{71.15}\\
\hline 
\end{tabular}%
}
\vspace{-5pt}
\caption{3D object detection results AP$_{50}$ at R$_{40}$ for pretraining on SemanticKITTI and fine-tuning on the full KITTI train set. Results are reported on the KITTI val set for easy (E), moderate (M), hard (H) difficulty levels.}
\label{tab:kitti3dod}
\end{table*}

We evaluate the transferability of our SSL model to a different downstream 3D scene perception task by fine-tuning on 3D object detection.   

\textbf{In-domain detection:} \Cref{tab:kitti3dod} shows our performance compared to SOTA published results \cite{tarl} for pretraining on SemanticKITTI and fine-tuning on the full KITTI train set. Our method outperforms \cite{nunes2022segcontrast} and \cite{tarl} on most difficulty levels even though \cite{tarl} has the advantage of temporal encoding. \Cref{tab:3dod} presents results for the model pretrained on Waymo and fine-tuned on 1\% Waymo, 5\% KITTI and 5\% nuScenes. On 1\% Waymo, we see significant improvement on all classes compared to training from scratch and on vehicle detection compared to SC. Our method, however, does not exceed SC's performance on small irregular objects such as pedestrians and cyclists. We hypothesize two reasons for this: 1) LiDAR pattern augmentation sparsifies the input point cloud, reducing the number of small clusters visible during pretraining. Hence, we see underfitting on small object classes. 2) Self-supervised bounding box targets tightly fit around only the visible parts of the object clusters, causing pretrained representations to overfit to imperfect box sizes. We suggest solutions to overcome these limitations in \cref{sec:future}.

\textbf{Cross-LiDAR detection:} On 5\% of KITTI, our method outperforms SC on all classes. Likewise, on 5\% nuScenes, PSA-SC performs better than SC on mAP over all classes. Since PSA-SC learns sensor-agnostic features via LiDAR pattern augmentation, it demonstrates better cross-LiDAR detection performance on all classes compared to SC. Overall, we see a strong domain transfer ability of our method to different beam patterns and a competitive performance on in-domain detection.  

\subsection{Ablation Studies}
In this section, we conduct ablation studies by pretraining SegContrast on Waymo with different design choices and components of our method and evaluating on semantic segmentation. We chose SegContrast as the baseline for our ablation studies since it is a stronger baseline than DepthContrast. \Cref{tab:ablations} presents the ablation results obtained by pretaining on 10 \% Waymo for 200 epochs and fine-tuning on 1\% Waymo, nuScenes and SemanticKITTI for 15 epochs. 
The first row shows the performance of the baseline SegContrast. 
Comparing PolarMix \cite{polarmix} (row 2) with Single Pattern LiDAR augmentation (row 3), the latter gives a larger improvement on all datasets indicating better cross-LiDAR generalizability. Hence, we choose single pattern augmentation for our final approach. 
Both Single Pattern LiDAR augmentation (row 3) and bounding box regression (row 4) separately improve performance over the baseline SegContrast.  However, combining both components (row 5) achieves the best results on all datasets. 

\begin{table}[tb]
\centering
\resizebox{\columnwidth}{!}{%
\begin{tabular}{c|ccc|c|c|c}
\toprule
Row \# & \multicolumn{1}{c|}{\begin{tabular}[c]{@{}c@{}} Polar\\ Mix\end{tabular}} 
& \multicolumn{1}{c|}{\begin{tabular}[c]{@{}c@{}}Single \\ Pattern\end{tabular}} 
& \multicolumn{1}{c|}{\begin{tabular}[c]{@{}c@{}}BB. \\ Reg.\end{tabular}}
& Waymo & nuScenes & SemKITTI  \\ 
\midrule
1 & & & & 40.17 & 32.64 & 37.63 \\
2 & \checkmark & & & 40.96 & 33.74 & 38.34 \\
3 & & \checkmark & & 43.44 & 34.88 & 39.09 \\
4 & & & \checkmark & 42.42 & 35.49 & 38.00 \\
5 & & \checkmark & \checkmark & \textbf{43.65} & \textbf{35.94} & \textbf{39.96} \\
\bottomrule
\end{tabular}%
}
\vspace{-5pt}
\caption{Ablation experiments done by fine-tuning on 1\% Waymo, nuScenes and SemanticKITTI labels for semantic segmentation (mIoU).}
\vspace{-5pt}
\label{tab:ablations}
\end{table}

\subsection{Latency Improvement}

One would expect that adding LiDAR pattern-augmentation and an additional regression pretext task would increase the pretraining time of the baseline SSL model. On our setup, pattern-augmentation takes 20 ms per point cloud on average. However, pattern-augmentation sparsifies point clouds, reducing the number of visible clusters, thus speeding up other augmentations and forward/backward passes. On average, PSA-SC takes 2.75 sec per pretraining iteration, whereas SC takes 4.1 sec. 
Hence, PSA-SSL reduces the training time of SC by 33\%. 

\subsection{Future Work} 
\label{sec:future}
To improve in-domain detection of small classes, we can augment the scene with more small clusters during pretraining. Small clusters close to the ground plane can be randomly sampled from other scenes in the batch. Further, one can refine the size of the bounding box targets by aggregating point clouds across multiple time steps. This would aggregate partial views of small objects, resulting in a full object cluster and accurate bounding box fitting. We leave the implementation of these extensions as future work.

\section{Conclusion}
In this paper, we identify and address fundamental limitations of contrastive-based SSL for LiDAR point clouds. We propose PSA-SSL, which extends existing contrastive methods with a self-supervised bounding box regression pretext task that constrains the point cloud features to predict object pose and size. Furthermore, we add a LiDAR beam pattern augmentation strategy to enable better generalizability of SSL features to different LiDAR sensors. 
Our approach is lightweight and model-agnostic.
Through experiments, we show significant performance improvements on 3D semantic segmentation and better cross-LiDAR transferability. We also show competitive performance on 3D object detection. 

\newpage

{
    \small
    \bibliographystyle{ieeenat_fullname}
    \bibliography{main}

\begin{thebibliography}{38}
\providecommand{\natexlab}[1]{#1}
\providecommand{\url}[1]{\texttt{#1}}
\expandafter\ifx\csname urlstyle\endcsname\relax
  \providecommand{\doi}[1]{doi: #1}\else
  \providecommand{\doi}{doi: \begingroup \urlstyle{rm}\Url}\fi

\bibitem[Behley et~al.(2019)Behley, Garbade, Milioto, Quenzel, Behnke, Stachniss, and Gall]{behley2019semantickitti}
Jens Behley, Martin Garbade, Andres Milioto, Jan Quenzel, Sven Behnke, Cyrill Stachniss, and Jurgen Gall.
\newblock Semantickitti: A dataset for semantic scene understanding of lidar sequences.
\newblock In \emph{Int. Conf. Comput. Vis.}, pages 9297--9307, 2019.

\bibitem[Boulch et~al.(2023)Boulch, Sautier, Michele, Puy, and Marlet]{ALSO}
Alexandre Boulch, Corentin Sautier, Björn Michele, Gilles Puy, and Renaud Marlet.
\newblock {ALSO}: Automotive lidar self-supervision by occupancy estimation.
\newblock In \emph{IEEE Conf. Comput. Vis. Pattern Recog.}, 2023.

\bibitem[Caesar et~al.(2019)Caesar, Bankiti, Lang, Vora, Liong, Xu, Krishnan, Pan, Baldan, and Beijbom]{nuscenes}
Holger Caesar, Varun Bankiti, Alex~H. Lang, Sourabh Vora, Venice~Erin Liong, Qiang Xu, Anush Krishnan, Yu Pan, Giancarlo Baldan, and Oscar Beijbom.
\newblock {NuScenes: A multimodal dataset for autonomous driving}.
\newblock \emph{arXiv preprint arXiv:1903.11027}, 2019.

\bibitem[Campello et~al.(2013)Campello, Moulavi, and Sander]{hdbscan}
Ricardo J. G.~B. Campello, Davoud Moulavi, and J{\"o}rg Sander.
\newblock Density-based clustering based on hierarchical density estimates.
\newblock In \emph{Pacific-Asia Conference on Knowledge Discovery and Data Mining}, 2013.

\bibitem[Caron et~al.(2020)Caron, Misra, Mairal, Goyal, Bojanowski, and Joulin]{caron2020unsupervised}
Mathilde Caron, Ishan Misra, Julien Mairal, Priya Goyal, Piotr Bojanowski, and Armand Joulin.
\newblock Unsupervised learning of visual features by contrasting cluster assignments.
\newblock In \emph{Adv. Neural Inform. Process. Syst.}, 2020.

\bibitem[Choy et~al.(2019)Choy, Gwak, and Savarese]{minkowski}
Christopher Choy, JunYoung Gwak, and Silvio Savarese.
\newblock 4d spatio-temporal convnets: Minkowski convolutional neural networks.
\newblock In \emph{IEEE Conf. Comput. Vis. Pattern Recog.}, pages 3075--3084, 2019.

\bibitem[Ester et~al.(1996)Ester, Kriegel, Sander, Xu, et~al.]{dbscan}
Martin Ester, Hans-Peter Kriegel, J{\"o}rg Sander, Xiaowei Xu, et~al.
\newblock A density-based algorithm for discovering clusters in large spatial databases with noise.
\newblock In \emph{Pacific-Asia Conference on Knowledge Discovery and Data Mining}, pages 226--231, 1996.

\bibitem[Geiger et~al.(2012)Geiger, Lenz, and Urtasun]{kitti}
Andreas Geiger, Philip Lenz, and Raquel Urtasun.
\newblock Are we ready for autonomous driving? the kitti vision benchmark suite.
\newblock In \emph{IEEE Conf. Comput. Vis. Pattern Recog.}, 2012.

\bibitem[Girshick(2015)]{fastrcnn}
Ross Girshick.
\newblock Fast r-cnn.
\newblock In \emph{Int. Conf. Comput. Vis.}, pages 1440--1448, 2015.

\bibitem[He et~al.(2020)He, Fan, Wu, Xie, and Girshick]{he2020momentum}
Kaiming He, Haoqi Fan, Yuxin Wu, Saining Xie, and Ross Girshick.
\newblock Momentum contrast for unsupervised visual representation learning.
\newblock In \emph{IEEE Conf. Comput. Vis. Pattern Recog.}, pages 9729--9738, 2020.

\bibitem[Hu et~al.(2023)Hu, Liu, and Hu]{hu2023density}
Qianjiang Hu, Daizong Liu, and Wei Hu.
\newblock Density-insensitive unsupervised domain adaption on 3d object detection.
\newblock In \emph{IEEE Conf. Comput. Vis. Pattern Recog.}, pages 17556--17566, 2023.

\bibitem[Krispel et~al.(2024)Krispel, Schinagl, Fruhwirth-Reisinger, Possegger, and Bischof]{maeli}
Georg Krispel, David Schinagl, Christian Fruhwirth-Reisinger, Horst Possegger, and Horst Bischof.
\newblock Maeli: Masked autoencoder for large-scale lidar point clouds.
\newblock In \emph{IEEE Conf. Comput. Vis. Pattern Recog.}, pages 3383--3392, 2024.

\bibitem[Lee et~al.(2022)Lee, Lim, and Myung]{patchworkpp}
Seungjae Lee, Hyungtae Lim, and Hyun Myung.
\newblock {Patchwork++: Fast and robust ground segmentation solving partial under-segmentation using 3D point cloud}.
\newblock In \emph{Proc. IEEE/RSJ Int. Conf. Intell. Robots Syst.}, pages 13276--13283, 2022.

\bibitem[Liang et~al.(2021)Liang, Jiang, Feng, Chen, Xu, Liang, Zhang, Li, and Van~Gool]{liang2021exploring}
Hanxue Liang, Chenhan Jiang, Dapeng Feng, Xin Chen, Hang Xu, Xiaodan Liang, Wei Zhang, Zhenguo Li, and Luc Van~Gool.
\newblock Exploring geometry-aware contrast and clustering harmonization for self-supervised 3d object detection.
\newblock In \emph{Int. Conf. Comput. Vis.}, pages 3293--3302, 2021.

\bibitem[Liu et~al.(2023)Liu, Kong, Cen, Chen, Zhang, Pan, Chen, and Liu]{seal}
Youquan Liu, Lingdong Kong, Jun Cen, Runnan Chen, Wenwei Zhang, Liang Pan, Kai Chen, and Ziwei Liu.
\newblock Segment any point cloud sequences by distilling vision foundation models.
\newblock In \emph{Adv. Neural Inform. Process. Syst.}, 2023.

\bibitem[Mahmoud et~al.(2023)Mahmoud, Hu, Kuai, Harakeh, Paull, and Waslander]{stslidr}
Anas Mahmoud, Jordan Hu, Tianshu Kuai, Ali Harakeh, Liam Paull, and Steven Waslander.
\newblock Self-supervised image-to-point distillation via semantically tolerant contrastive loss.
\newblock In \emph{IEEE Conf. Comput. Vis. Pattern Recog.}, 2023.

\bibitem[Min et~al.(2023)Min, Xiao, Zhao, Nie, and Dai]{occmae}
Chen Min, Liang Xiao, Dawei Zhao, Yiming Nie, and Bin Dai.
\newblock Occupancy-mae: Self-supervised pre-training large-scale lidar point clouds with masked occupancy autoencoders.
\newblock \emph{IEEE Transactions on Intelligent Vehicles}, 2023.

\bibitem[Nunes et~al.(2022)Nunes, Marcuzzi, Chen, Behley, and Stachniss]{nunes2022segcontrast}
Lucas Nunes, Rodrigo Marcuzzi, Xieyuanli Chen, Jens Behley, and Cyrill Stachniss.
\newblock Segcontrast: 3d point cloud feature representation learning through self-supervised segment discrimination.
\newblock \emph{IEEE Robotics and Automation Letters}, 7\penalty0 (2):\penalty0 2116--2123, 2022.

\bibitem[Nunes et~al.(2023)Nunes, Wiesmann, Marcuzzi, Chen, Behley, and Stachniss]{tarl}
Lucas Nunes, Louis Wiesmann, Rodrigo Marcuzzi, Xieyuanli Chen, Jens Behley, and Cyrill Stachniss.
\newblock {Temporal Consistent 3D LiDAR Representation Learning for Semantic Perception in Autonomous Driving}.
\newblock In \emph{IEEE Conf. Comput. Vis. Pattern Recog.}, 2023.

\bibitem[Oord et~al.(2018)Oord, Li, and Vinyals]{infonce}
Aaron van~den Oord, Yazhe Li, and Oriol Vinyals.
\newblock Representation learning with contrastive predictive coding.
\newblock \emph{arXiv preprint arXiv:1807.03748}, 2018.

\bibitem[Ryu et~al.(2023)Ryu, Hwang, and Park]{lidomaug}
Kwonyoung Ryu, Soonmin Hwang, and Jaesik Park.
\newblock Instant domain augmentation for lidar semantic segmentation.
\newblock In \emph{IEEE Conf. Comput. Vis. Pattern Recog.}, pages 9350--9360, 2023.

\bibitem[Sautier et~al.(2022)Sautier, Puy, Gidaris, Boulch, Bursuc, and Marlet]{slidr}
Corentin Sautier, Gilles Puy, Spyros Gidaris, Alexandre Boulch, Andrei Bursuc, and Renaud Marlet.
\newblock Image-to-lidar self-supervised distillation for autonomous driving data.
\newblock In \emph{IEEE Conf. Comput. Vis. Pattern Recog.}, pages 9891--9901, 2022.

\bibitem[Sautier et~al.(2024)Sautier, Puy, Boulch, Marlet, and Lepetit]{sautier2024bevcontrast}
Corentin Sautier, Gilles Puy, Alexandre Boulch, Renaud Marlet, and Vincent Lepetit.
\newblock Bevcontrast: Self-supervision in bev space for automotive lidar point clouds.
\newblock In \emph{Int. Conf. 3D Vision}, pages 559--568. IEEE, 2024.

\bibitem[Shi et~al.(2019)Shi, Wang, Li, et~al.]{pointrcnn}
S Shi, X Wang, H~Pointrcnn Li, et~al.
\newblock 3d object proposal generation and detection from point cloud.
\newblock In \emph{IEEE Conf. Comput. Vis. Pattern Recog.}, 2019.

\bibitem[Shi et~al.(2021)Shi, Wang, Shi, Wang, and Li]{parta2}
Shaoshuai Shi, Zhe Wang, Jianping Shi, Xiaogang Wang, and Hongsheng Li.
\newblock From points to parts: 3d object detection from point cloud with part-aware and part-aggregation network.
\newblock \emph{IEEE Trans. Pattern Anal. Mach. Intell.}, 43\penalty0 (8):\penalty0 2647--2664, 2021.

\bibitem[Shi and Rajkumar(2022)]{shi2022self}
Weijing Shi and Ragunathan~Raj Rajkumar.
\newblock Self-supervised pretraining for point cloud object detection in autonomous driving.
\newblock In \emph{2022 IEEE 25th International Conference on Intelligent Transportation Systems (ITSC)}, pages 4341--4348. IEEE, 2022.

\bibitem[Sun et~al.(2020)Sun, Kretzschmar, Dotiwalla, Chouard, Patnaik, Tsui, Guo, Zhou, Chai, Caine, et~al.]{waymo}
Pei Sun, Henrik Kretzschmar, Xerxes Dotiwalla, Aurelien Chouard, Vijaysai Patnaik, Paul Tsui, James Guo, Yin Zhou, Yuning Chai, Benjamin Caine, et~al.
\newblock Scalability in perception for autonomous driving: Waymo open dataset.
\newblock In \emph{IEEE Conf. Comput. Vis. Pattern Recog.}, pages 2446--2454, 2020.

\bibitem[Xiao et~al.(2022)Xiao, Huang, Guan, Cui, Lu, and Shao]{polarmix}
Aoran Xiao, Jiaxing Huang, Dayan Guan, Kaiwen Cui, Shijian Lu, and Ling Shao.
\newblock Polarmix: A general data augmentation technique for lidar point clouds.
\newblock \emph{arXiv preprint arXiv:2208.00223}, 2022.

\bibitem[Xie et~al.(2020)Xie, Gu, Guo, Qi, Guibas, and Litany]{xie2020pointcontrast}
Saining Xie, Jiatao Gu, Demi Guo, Charles~R Qi, Leonidas Guibas, and Or Litany.
\newblock Pointcontrast: Unsupervised pre-training for 3d point cloud understanding.
\newblock In \emph{Eur. Conf. Comput. Vis.}, pages 574--591. Springer, 2020.

\bibitem[Yan et~al.(2023)Yan, Chen, Zhang, Yuan, Cai, Shi, Shao, Yan, Luo, and Qiao]{yan2023spot}
Xiangchao Yan, Runjian Chen, Bo Zhang, Jiakang Yuan, Xinyu Cai, Botian Shi, Wenqi Shao, Junchi Yan, Ping Luo, and Yu Qiao.
\newblock Spot: Scalable 3d pre-training via occupancy prediction for autonomous driving.
\newblock \emph{arXiv preprint arXiv:2309.10527}, 2023.

\bibitem[Yan et~al.(2024)Yan, Zhang, Cai, Guo, Qiu, Gao, Zhou, Zhao, Jin, Gao, et~al.]{yan2024forging}
Xu Yan, Haiming Zhang, Yingjie Cai, Jingming Guo, Weichao Qiu, Bin Gao, Kaiqiang Zhou, Yue Zhao, Huan Jin, Jiantao Gao, et~al.
\newblock Forging vision foundation models for autonomous driving: Challenges, methodologies, and opportunities.
\newblock \emph{arXiv preprint arXiv:2401.08045}, 2024.

\bibitem[Yan et~al.(2018)Yan, Mao, and Li]{yan2018second}
Yan Yan, Yuxing Mao, and Bo Li.
\newblock Second: Sparsely embedded convolutional detection.
\newblock \emph{Sensors}, 18\penalty0 (10):\penalty0 3337, 2018.

\bibitem[Yang et~al.(2024)Yang, Zhang, Huang, Wu, Zhu, He, Tang, Zhao, Qiu, Lin, et~al.]{unipad}
Honghui Yang, Sha Zhang, Di Huang, Xiaoyang Wu, Haoyi Zhu, Tong He, Shixiang Tang, Hengshuang Zhao, Qibo Qiu, Binbin Lin, et~al.
\newblock Unipad: A universal pre-training paradigm for autonomous driving.
\newblock In \emph{IEEE Conf. Comput. Vis. Pattern Recog.}, pages 15238--15250, 2024.

\bibitem[Yin et~al.(2022)Yin, Zhou, Zhang, Fang, Xu, Shen, and Wang]{yin2022proposalcontrast}
Junbo Yin, Dingfu Zhou, Liangjun Zhang, Jin Fang, Cheng-Zhong Xu, Jianbing Shen, and Wenguan Wang.
\newblock Proposalcontrast: Unsupervised pre-training for lidar-based 3d object detection.
\newblock In \emph{Eur. Conf. Comput. Vis.}, pages 17--33. Springer, 2022.

\bibitem[You et~al.(2022)You, Luo, Phoo, Chao, Sun, Hariharan, Campbell, and Weinberger]{modest}
Yurong You, Katie Luo, Cheng~Perng Phoo, Wei-Lun Chao, Wen Sun, Bharath Hariharan, Mark Campbell, and Kilian~Q Weinberger.
\newblock Learning to detect mobile objects from lidar scans without labels.
\newblock In \emph{IEEE Conf. Comput. Vis. Pattern Recog.}, pages 1130--1140, 2022.

\bibitem[Zhang et~al.(2023)Zhang, Yang, Xiong, Casas, Yang, Ren, and Urtasun]{oyster}
Lunjun Zhang, Anqi~Joyce Yang, Yuwen Xiong, Sergio Casas, Bin Yang, Mengye Ren, and Raquel Urtasun.
\newblock Towards unsupervised object detection from lidar point clouds.
\newblock In \emph{IEEE Conf. Comput. Vis. Pattern Recog.}, pages 9317--9328, 2023.

\bibitem[Zhang et~al.(2017)Zhang, Xu, Dong, and Dolan]{lshape}
Xiao Zhang, Wenda Xu, Chiyu Dong, and John~M. Dolan.
\newblock Efficient l-shape fitting for vehicle detection using laser scanners.
\newblock In \emph{2017 IEEE Intelligent Vehicles Symposium (IV)}, pages 54--59, 2017.

\bibitem[Zhang et~al.(2021)Zhang, Girdhar, Joulin, and Misra]{Zhang_2021_depthcontrast}
Zaiwei Zhang, Rohit Girdhar, Armand Joulin, and Ishan Misra.
\newblock Self-supervised pretraining of 3d features on any point-cloud.
\newblock In \emph{Int. Conf. Comput. Vis.}, pages 10252--10263, 2021.

\end{thebibliography}
}

% WARNING: do not forget to delete the supplementary pages from your submission 
\clearpage
\setcounter{page}{1}
\maketitlesupplementary

\subsection{Sensitivity to hyperparameters}
In this section, we study the sensitivity of PSA-SSL to LiDAR pattern augmentation probabilities and clustering $\epsilon$. All models, unless mentioned otherwise, are pretrained on 10\% Waymo for 30 epochs and fine-tuned on 1\% of the labels for 15 epochs.

\subsubsection{Sensitivity to pattern augmentation probabilities}
\label{sec:sens_p32}
The hyperparameters for LiDAR pattern augmentation are the probabilities with which we randomly sample different LiDAR configurations (`v32', `v64', `o64'). 
\cref{tab:pattern} shows semantic segmentation performance when the probabilities for randomly sampling LiDAR configurations are changed. For simplicity, we increase the probability for `v32' from 0.3 to 0.6 and equally divide 1-prob(v32) between `v64' and `o64'. As expected, higher probabilities for v64 and o64 (\ie 
 prob(v32) $\in [0.3, 0.4]$) give better performance on dense LiDAR (Waymo and SemanticKITTI). Increasing v32 probability from 0.3 to 0.4 and 0.5 improves performance on nuScenes, but beyond 0.5, the performance on all datasets decreases, possibly due to a decrease in visible clusters during pretraining. Increasing the number of pretraining epochs from 30 to 200 allows the prob(v32)=0.6 to catch up to the performance of prob(v32)=0.4. Ablation studies in \cref{tab:ablations} also show that with prob(v32)=0.6, SegContrast benefits from LiDAR pattern augmentation across all datasets (see row 1 vs row 3). 
Hence, we conclude that our method is robust to the pattern augmentation probabilities. 

\begin{table}[h!]
\centering
\resizebox{\columnwidth}{!}{%
\begin{tabular}{ccccc}
\toprule
Epochs & Prob. (v32) &  1\% Waymo &  1\% nuScenes & 1\% SemKITTI \\ 
\midrule
\multirow{5}{*}{\begin{tabular}{c}
    Pretrain=30 \\
    Fine-tune=15
\end{tabular}} & No pretraining & 23.24 & 23.59 & 19.51 \\
& 0.3 & \underline{37.8} & 32.76 & \textbf{36.06} \\
& 0.4 & \textbf{39.79} & \textbf{33.80} & \underline{34.96} \\
& 0.5 & 37.06 & \underline{33.04} & 34.46 \\
& 0.6 & 36.90 & 32.83 & 34.54 \\ \hline
\multirow{2}{*}{\begin{tabular}{c}
    Pretrain=200 \\
    Fine-tune=100
\end{tabular}} & 0.4 & \textbf{54.61} & 37.56 & 51.71 \\
& 0.6 & 54.36 & \textbf{37.89} & \textbf{52.11} \\ 
\bottomrule 
\end{tabular}%
}
\vspace{-5pt}
\caption{Semantic Segmentation (mIoU) sensitivity to LiDAR pattern augmentation probabilities.}
\vspace{-10pt}
\label{tab:pattern}
\end{table}

\subsubsection{Sensitivity to clustering epsilon}
\label{sec:sens_eps}
Incorrectly setting the clustering $\epsilon$ may lead to over or under-segmentation for SegContrast, which can also affect the bounding box regression pretext task, resulting in lower semantic segmentation performance. Since SegContrast \cite{nunes2022segcontrast} does not study the effect of clustering $\epsilon$ on their method, we analyse the impact of changing $\epsilon$ on PSA-SC's segmentation performance. \Cref{tab:sens_eps} shows that the performance of PSA-SC does not drop dramatically in the range of $\epsilon \in [0.2, 0.3]$. Hence, there is a range of $\epsilon$, for which the performance is robust. We also note that PSA-SC outperforms training from scratch at all clustering $\epsilon$ and outperforms its baseline SC at the best $\epsilon=0.2$. From the table, we can see that $\epsilon=0.2$ is closest to the ideal $\epsilon$, which matches the value we set by visually tuning on a few frames. Since visually tuning $\epsilon$ is trivial, we do not consider it as a limitation of our method. 

\begin{table}[h!]
\centering
\resizebox{\columnwidth}{!}{%
\begin{tabular}{c|cc|cccc}
\toprule
Method &  No Pretraining & SC & \multicolumn{4}{c}{PSA-SC (Ours) }\\ 
$\epsilon$ & - & 0.2 & 0.1 & 0.2 & 0.3 & 0.4 \\
\midrule
mIoU & 23.24 & 38.50 & 36.71  & \textbf{39.79} & 38.22 & 36.06 \\
\hline 
\end{tabular}%
}
\vspace{-5pt}
\caption{Semantic Segmentation (mIoU) sensitivity to clustering $\epsilon$. Models are fine-tuned on 1\% Waymo.}
\label{tab:sens_eps}
\vspace{-10pt}
\end{table}

\section{Qualitative Results}

\textbf{\begin{figure}[tb]
  \centering
  \includegraphics[width=0.9\columnwidth]{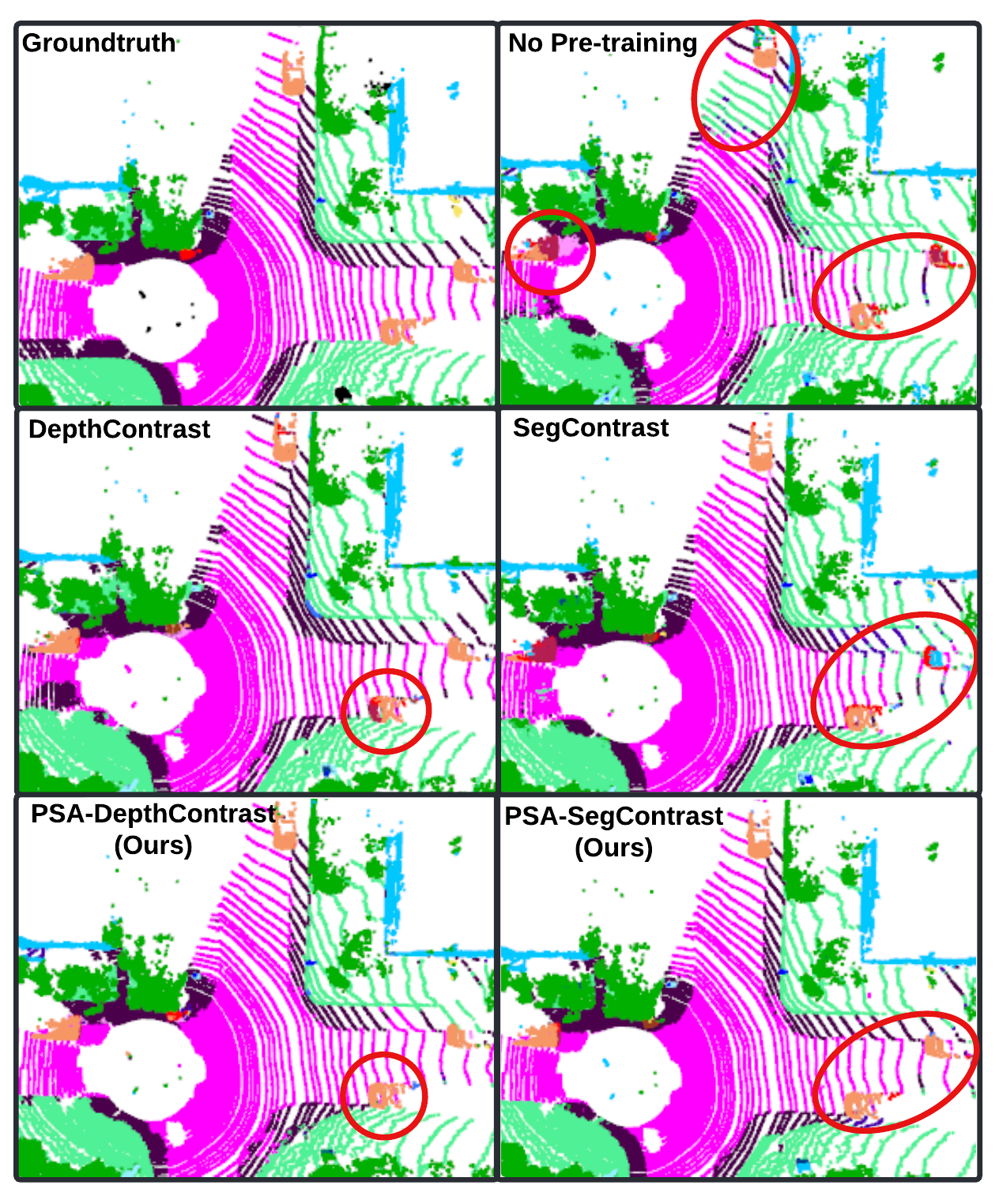}
  % \vspace{-20pt}
  \caption{Comparison of qualitative semantic segmentation results of PSA-DepthContrast and PSA-SegContrast against their original baselines  \cite{Zhang_2021_depthcontrast, nunes2022segcontrast}  on SemanticKITTI validation scan.}
  \label{fig:qualitative}
  % \vspace{-10pt}
\end{figure}}

\vspace{-10pt}
\Cref{fig:qualitative} shows a qualitative comparison of semantic segmentation between our approach and baselines. As expected, training from randomly initialized backbone errs on vehicles as well as leaks labels to neighbouring stuff classes. DepthContrast and SegContrast exhibit label confusion on vehicles but improve on stuff classes. Incorporating PSA, significantly minimizes the label confusion on vehicles, indicating the ability to learn features that can capture full extent of objects in the scene, resulting in better geomtery-aware class discrimination.

\end{document}